\documentclass[letterpaper]{article} 
\usepackage{aaai23}  
\usepackage{times}  
\usepackage{helvet}  
\usepackage{courier}  
\usepackage[hyphens]{url}  
\usepackage{graphicx} 
\usepackage{natbib}  
\usepackage{caption} 
\usepackage{algorithm}
\usepackage{algorithmic}

\usepackage{newfloat}
\usepackage{listings}
\DeclareCaptionStyle{ruled}{labelfont=normalfont,labelsep=colon,strut=off} 
\floatstyle{ruled}
\newfloat{listing}{tb}{lst}{}
\floatname{listing}{Listing}
\usepackage{multirow}
\usepackage{svg}
\usepackage{longtable}
\usepackage{amsmath}

\title{Improving Long-Horizon Imitation through Instruction Prediction}
\author {
    Joey Hejna\textsuperscript{\rm 1},
    Pieter Abbeel\textsuperscript{\rm 2},
    Lerrel Pinto\textsuperscript{\rm 3}
}
\affiliations {
    \textsuperscript{\rm 1} Stanford University\\
    \textsuperscript{\rm 2} University of California, Berkeley\\
    \textsuperscript{\rm 3} New York University \\
    jhejna@cs.stanford.edu, pabbeel@berkeley.edu, lerrel@cs.nyu.edu
}

\usepackage{bibentry}

\begin{document}

\maketitle

\begin{abstract}
Complex, long-horizon planning and its combinatorial nature pose steep challenges for learning-based agents. Difficulties in such settings are exacerbated in low data regimes where over-fitting stifles generalization and compounding errors hurt accuracy. In this work, we explore the use of an often unused source of auxiliary supervision: language. Inspired by recent advances in transformer-based models, we train agents with an instruction prediction loss that encourages learning temporally extended representations that operate at a high level of abstraction. Concretely, we demonstrate that instruction modeling significantly improves performance in planning environments when training with a limited number of demonstrations on the BabyAI and Crafter benchmarks. In further analysis we find that instruction modeling is most important for tasks that require complex reasoning, while understandably offering smaller gains in environments that require simple plans. More details and code can be found at \url{https://github.com/jhejna/instruction-prediction}.

\end{abstract}

\section{Introduction}

Intelligent agents ought to be able to complete complex, long horizon tasks and generalize to new scenarios. Unfortunately, policies learned by modern deep-learning techniques often struggle to acquire either of these abilities. This is particularly true in planning regimes where multiple, complex, steps must be completed correctly in sequence to complete a task. Realistic constraints, such as partial observability, the underspecification of goals, or the sparse reward nature of many planning problems make learning even harder. Reinforcement learning approaches often struggle to effectively learn policies and require billions of environment interactions to produce effective solutions \cite{wijmans2019dd, parisotto2020stabilizing}. Imitation learning is an alternative approach based on learning from expert data, but can still require millions of demonstrations to learn effective planners \cite{babyai_iclr19}. Such high data constraints make learning difficult and expensive. 

Unfortunately the aforementioned issues with behavior learning are only exacerbated in the low data regime. First, with limited training data agents are less likely to act perfectly at each environment step, leading to small errors that compound overtime in the offline setting. Ultimately, this leads to sub-par performance over long horizons that can usually only be improved by carefully collecting additional expert data \cite{ross2011reduction}. Second, deep-learning based policies are more likely to overfit small training datasets, making them unable to generalize to new test-time scenarios.
On the other hand, humans have the remarkable ability to interpolate previous knowledge and solve unseen long-horizon tasks. After observing an environment, we might deduce plan or sequence of the steps to follow to complete our objective. However, imitation learning agents are not required to construct plans by default -- they are usually trained to only output action sequences given seen observations. This begs the question: how can we make agents reason better in long-horizon tasks? 


An attractive solution lies in language instructions, the same medium humans use for mental planning \cite{gleitman2005language}. \textcolor{black}{Several prior works directly provide agents with language instructions to follow \cite{mattersim, ALFRED20, chen2019touchdown}. Unfortunately, such approaches require the specification of exhaustive instructions at test time for systems to function. A truly intelligent agent ought to be able to devise its own plan and execute it, with only a handful of demonstrations.} We propose improving policy learning in the low-data regime by having agents predict planning instructions in addition to their immediate next action. \textcolor{black}{As we do not input instructions to the policy, we can plan without their specification at test time. Though prior works have used hierarchical structures that generate their own instructions to condition on \cite{chen2021ask,hu2019hierarchical,jiang2019language}, we surprisingly find that just predicting language instructions is in itself a powerful objective to learn good representations for planning.} Teaching agents to output language instructions for completing tasks has two concrete benefits. First, it forces them to learn at a higher level of abstraction where generalization is easier. Second, by outputting multi-step instructions agents explicitly consider the future. Practically, we teach agents to output instructions by adding an auxiliary instruction prediction network to transformer-based policy networks, as in seq2seq translation \cite{attention2017vaswani}. Our approach can be interpreted as translating observations or trajectories into instructions.

We test our representation learning method in limited data settings and combinatorially complex enviornments. 
We find that in many settings higher performance can be attained by relabeling existing demonstrations with language instructions instead of collecting new ones, creating a new, scalable type of data collection for practitioners. Furthermore, our method is conceptually simple and easy to implement. This work is the first to show that direct representation learning with language can accelerate imitation learning.

To summarize, our contributions are as follows. First, we introduce a method for training transformer based planning networks on paired demonstration and instruction data via an auxiliary instruction prediction loss. Second, we test our objective in long-horizon planning based environments with limited data and find that it substantially outperforms contemporary approaches. Finally, we analyze the scenarios in which predicting instructions provides fruitful training signal, concluding that instruction modeling is a valuable objective when tasks are sufficiently complex. 

\section{Related Work}

Language in the context of policy learning has been heavily studied \cite{luketina2019survey}, usually to communicate a task objective. Uniquely, we use natural language instructions to aid in learning via an auxiliary objective. Here we survey the most relevant works to our approach.


\textbf{Language Goals.}
Language offers a natural medium to communicate goals to intelligent agents. As such, several prior work have focused on learning language goal conditioned policies, particularly for robotics \cite{nair2021learning, stepputtis2020language, kanu2020following, hill2020human, akakzia2021grounding, goyal2021zero, shridhar2021cliport}, or for games \cite{babyai_iclr19, chaplot2018gated, hermann2017grounded}, and sometimes even with hindsight relabeling \cite{chan2019actrce, cideron2019self}. Others in the area of inverse reinforcement learning use language to specify reward functions \cite{fu2019language, bahdanau2018learning, williams2018learning} or shape them \cite{mirchandani2021ella, goyal2019using}. These works use language to give humans an easy way to specify the desired goal conditions of an environment. Unlike these works, we use language instructions that dictate the steps to reach a desired goal condition instead of just using language goals that specify the desired state. Other works, particularly in the visual navigation space, provide agents with step-by-step instructions similar to those we use, sometimes in addition to language goals. \citet{mattersim,fried2018speaker, chen2019touchdown, krantz2020beyond, chen2021topological, chen2019touchdown, zang2018translating} condition policies on step-by-step language instructions for visual navigation, while \citet{ALFRED20, pashevich2021episodic, ALFWorld20} use instructions for household tasks. Many of these benchmarks ask agents to simply follow instructions, like ``turn right at the end of the hallway'' instead of achieving overarching goals like ``go to the kitchen''. Critically unlike our method, these approaches require laboriously providing the agent with step-by-step instructions at test-time. By using instructions for representation learning instead of policy inputs, we additionally avoid needing to label entire datasets with language instructions and can train on partially labeled datasets. \textcolor{black}{Other proposed environments \cite{wang2021grounding} assess understanding by prompting agents with necessary information about task dynamics, precluding the removal of text-prompting at test time.}

\textbf{Language and Hierarchical Learning.}
Instead of directly using instructions as policy inputs, other works use language instructions as an intermediary representations for hierarchical policies. Usually, a high-level planner outputs language instructions for a low-level executor to follow. \citet{andreas2017modular} and \citet{oh2017zero} provide agents with hand-designed high-level language instructions or policy ``sketches''. Again unlike our method such approaches require  instruction labels for every training task and for every new task at test-time. \citet{jiang2019language} and \citet{shu2017hierarchical} provide interactive language labels to agents to train hierarchical policies with reinforcement learning. In the imitation learning setting, \citet{hu2019hierarchical} learn a hierarchical policy using behavior cloning for a strategy game. Unlike the planning problems we consider, their environment has no oracle solution and does not consider generalization to unseen tasks. Most related to our work, \citet{chen2021ask} use latent representations from a learned high-level instruction predictor to aid a low-level policy. However, unlike \citet{chen2021ask}, we learn latent representations that can predict instructions, but do not explicitly condition on them at test-time. While these hierarchical approaches have shown promise, the quality of learned policies is inherently limited by the amount of language data available for training. Even with a perfect low-level policy, inaccurate  languages commands will yield poor overall performance. For example, a small mis-specification in a subgoal, like changing ``blue door'' to ``red door'' would likely cause complete policy failure. This is not an issue for our loss-based approach, as our instruction prediction network can be detached from the policy. As previously mentioned, this structure lets us learn on a mix of instruction annotated and unannotated data, letting it more easily scale than hierarchical approaches particularly in data-limited scenarios. Other approaches in robotics similar to hierarchy use high-level discrete action labels alongside demonstrations to learn planning grammars \citep{edmonds2017feeling, edmonds2019tale}. While different in flavor than our approach, such methods also share similar data limitations to the hierarchical methods preivously discussed.

\textbf{Auxiliary Objectives.}
The use of auxiliary objectives in policy learing has been extensively studied. Though to our knowledge no prior have used instructions, auxiliary objectives in general have been found to aid policy learning \cite{jaderberg2016reinforcement}. \citet{laskin2020curl} and \citet{stooke2021decoupling} demonstrated the success of contrastive auxiliary objectives in robotic reinforcement learning domains. \citet{schwarzer2020data} and \citet{anand2019unsupervised} did the same in the Atari game-playing environments. We were inspired by their effectiveness. Additionally, works like \citet{andreas2018learning} have previously used language question and answering for representation leaning in visual domains.

\textbf{Transformers.} Our approach is based on several innovations involving transformer networks. \citet{attention2017vaswani} previously showed state of the art results in machine translation using transformers. While the application of transformers has extended to behavior learning \cite{zambaldi2018deep, parisotto2020stabilizing, chen2021decision}, prior works in the area have not leveraged the transformer decoder. Closest to our domain, \citet{lin2021vx2text} generate captions from video. The architecture of our policy networks take inspiration from recent works adapting transformers to mediums beyond text, namely in vision \cite{image2021kolesnikov} and offline reinforcement learning \cite{chen2021decision}.

\section{Method}
In this section we formally describe imitation learning with instruction prediction, then detail our implementation for both Markovian and non-Markovian environments. 

\subsection{Problem Setup}
The standard learning from demonstrations setup assumes access to a dataset of expert trajectory sequences containing paired observations and actions $o_1, a_1, o_2, a_2, ... , o_T, a_T$. The goal of imitation learning is to learn a policy $\pi(a_t|\cdot)$ that predicts the correct actions an agent should take. In our work we consider both Markovian and partially observed non-Markovian settings. In the non-Markovian observed case, policies are given access to previous observations in order to infer state information \cite{kaelbling1998planning}, and we denote the policy as $\pi(a_t|o_1, ... o_t)$. In the Markovian setting this is unnecessary, and the policy is simply $\pi(a_t|o_t)$. In imitation learning it is common for policies to be goal conditioned, or even conditioned on language goals as is the case in our experiments. In goal conditioned settings an encoding of the desired task or goal $g$ is additional input to the policy. As our approach works with or without goal conditioning we omit it from the rest of this section for brevity. A standard imitation learning technique is behavior cloning, which in discrete domains maximizes the likelihood of the actions in the dataset using a negative log likelihood objective, $\mathcal{L}_{\textrm{action}} = - \sum_{t} \log \pi(a_t|\cdot)$.


In this work, we assume access to oracle language instructions that tell an agent how it should complete a task to provide useful training signal. As mentioned in Section 2, for the purposes of our method we distinguish goals from instructions. Language goals describe the desired final state of the environment environment, specifying \textit{what} to do, whereas language instructions communicate \textit{how} an agent should reach the desired state in a step-by-step manner. Each trajectory may have several language instructions $x^{(1)}, x^{(2)}, ..., x^{(n)}$ corresponding to each of the $n$ different steps to reach the desired goal configuration. For example, a language instruction like ``open the door'' only applies to the part of the demonstration before the agent opens the door and after it completes the last instruction. The $i$-th instruction $x^{(i)}$ thus corresponds to an interval $[T_i, T_{i+1})$ where $T_i$ marks the time the instruction was given and $T_{i+1}$ denotes the start of the next instruction. A depiction of an example instruction sequence can be found in Figure \ref{fig:env}. While language instructions are an additional data requirement, they can be cheap to obtain, particularly in scenarios where demonstrations are expensive to collect. \textcolor{black}{If demonstrations are collected in the real world, providing simultaneous instruction annotations for a demonstration is likely less labor-intensive than collecting an additional demonstration. Moreover, humans can easily re-label existing demonstrations with instructions. Video data could easily be captioned with voice-over. Similar statements can be made for simulators -- if one can code an oracle policy, instructions are likely easy to generate along the way. These modifications are easy for simulators with planning stacks, as in BabyAI.} Moreover, we focus on low to medium data regime, where the cost of setting up an environment and collecting more demonstrations is likely be higher than annotating an existing small set of demonstrations. Next, we describe how we train agents to predict language instructions to aid in imitation learning.




\begin{figure*}[]
\centering
\includegraphics[width=0.8\textwidth]{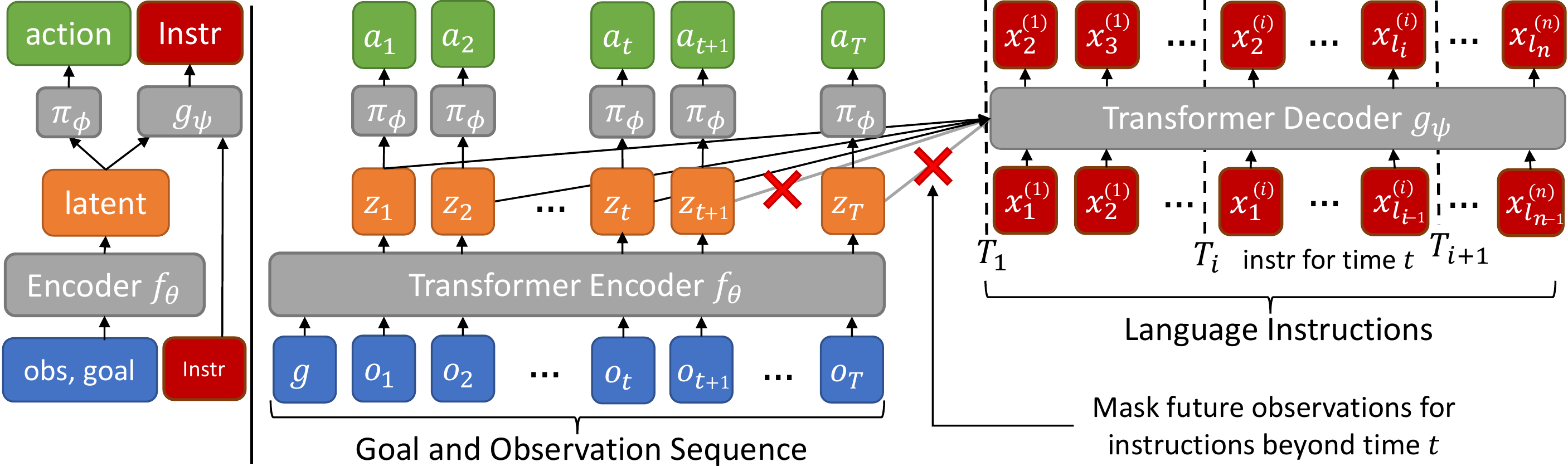}
\caption{The left diagram depicts the general model architecture used for our approach. Notice how the policy and encoder can be completed separated from the instruction component for mixed-data training or inference. The diagram on the right depicts its implementation for the partial observed environments using a GPT-like transformer encoder. The diagram shows our masking scheme at episode step $t$: latent vectors from beyond time $t$ are masked from the language decoder.}
\label{fig:arch}
\end{figure*}

\subsection{Instruction Prediction for Imitation Learning}
\label{sec:instr_pred}
The central hypothesis of this work is that predicting language instructions will force agents to learn representations beneficial for long-horizon planning. In our framework, we first construct an observation encoder $f_\theta$ that produces latent representations $z$. As in behavior cloning we predict actions from latents $z$ using a policy network $\pi_\phi$, but we additionally use a language decoder $g_\psi$ to predict the current instruction from $z$. Our general setup is shown in left half of Figure \ref{fig:arch}. We consider both non-Markovian environments, where sequences of observations must be provided to the model so it can infer the underlying state. In the non-Markovian setting, the encoder produces $z_1, ..., z_t = f_\theta(o_1, ...,o_t)$, the policy is $\pi_\phi(a_t|z_1, ..., z_t)$, and the language model is $g_\psi(x^{(i)}|z_1, ..., z_{T_i - 1})$. For standard fully observed Markovian environments, conditioning on past observations is unnecessary and the encoder, policy, and language decoder can be written as $z_t = f(o_t)$, $\pi_\phi(a_t|z_t)$, and $g_\psi(x^{(i)}|z_t)$ respectively. As is common in natural language processing, we treat each language instruction $x^{(i)}$ as a sequence of multiple text tokens $x^{(i)}_1,x^{(i)}_2, ..., x^{(i)}_{l_i}$ where $l_i$ is the length of the $i$-th instruction. The decoder is trained using the standard language modeling loss. We construct our total imitation learning objective by maximizing the log-likelihood of both the action and instruction data. For a given trajectory in the non-Markovian case, this is written as follows
\begin{align}
    \mathcal{L} &= -\sum_{t=1}^T \log \pi_\phi(a_t | z_1, ..., z_t) \\ &- \lambda \sum_{i=1}^n \sum_{j=1}^{l_i} \log g_\psi(x^{(i)}_j | x_1^{(i)}, ..., x^{(i)}_{j-1}, z_1, ..., z_{T_i - 1}) \nonumber
\end{align}

where latent representations $z$ are all produced by the shared encoder $f_\theta$. The MDP case is formulated by removing past conditioning on $z_1, ..., z_{t-1}$. The first term of the loss is the standard classification loss used for behavior cloning in discrete domains. The second term of the loss corresponds to the negative log-likelihood of the language instructions. We index the language loss by instructions via the first sum. The second sum over token log likelihoods is from the standard auto-regressive language modeling framework, where the likelihood of an instruction is the product of the conditional probabilities $p(x^{(i)}) = \prod_{j=1}^{l_i} p(x^{(i)}_j |x^{(i)}_1, ... x^{(i)}_{j-1})$. Note that when predicting the likelihood of language tokens for instruction $i$, the model can only condition on latents up to $z_{T_i -1}$. This ensures that we compute the likelihood of instruction $i$ using only observations during or before its execution. Finally, $\lambda$ is a weighting coefficient that trades off the importance of instruction prediction and action modeling. During training, we optimize all parameters $\phi, \psi$, and $\theta$ jointly, meaning that gradients from both behavior cloning and language prediction are propagated to the encoder weights $\theta$. In some of our experiments we test additional learning objectives which are also trained on top of the same latent representations $z$ as is standard in the literature \cite{jaderberg2016reinforcement}.


Though our method is general to any network architecture, we train transformer based policies since they have been shown to be extremely effective at natural language processing tasks \cite{attention2017vaswani} and carry a good inductive bias for combinatorial planning problems \cite{zambaldi2018deep}. For details on the transformer architectures we use, we defer to \cite{image2021kolesnikov,chen2021decision}. In the following sections we describe our transformer-based models for both Markovian and non-Markovian settings. 

\textbf{Non-Markovian Settings.} For environments that are non-Markovian or partially observed we use a transformer based sequence model as our policy network, similar to those employed in \cite{chen2021decision}. We operate in the entire sequence at once: $z_1, ... z_T = f_\theta(o_1, ..., o_T)$. Causal masking similar to that in \cite{radford2018improving} ensures that at time $t$ the representation $z_t$ only depends on current and previous observations $o_1, ... o_t$. The same policy network $\pi_\phi(a_t|z_t)$ is applied to each latent to produce actions for each timestep. The language decoder $g_\psi$ is also a transformer model and employs both causal attention masks to the language inputs and cross attention masks to the latents. Causal-self attention masks on the language inputs enforce the auto-regressive modeling of the instruction tokens. Cross attention masks to the latent representations ensure that predictions for the $i$th instruction cannot attend to latents from timesteps after its execution as is depicted by the red ``x''s in Figure \ref{fig:arch}. This forces language prediction during training to mirror test-time as the agent cannot use future information to predict the instruction.  


\textbf{Markovian settings.} For environments that are fully Markovian, we use only the most recent observation $o_t$. As sequence modeling is unnecessary, we use a transformer to encode individual states, leveraging their success in combinatorial environments \citep{zambaldi2018deep}. Specifically, we a Vision Transformer architecture \citep{image2021kolesnikov} that predict actions only for a single timestep. Observations are preprocessed into tokens and prepended with a special CLS token: $o_t \rightarrow \textrm{CLS}, o_{t,1}, o_{t,2}, o_{t,3}, ...$. As we do not input future observations, the transformer encoder uses full unmasked self attention. At the end of the network we take the latent representation corresponding to the CLS token and use it to predict the action $\pi_\phi(a_t | z_{t, \textrm{CLS}})$. We use all latent tokens $z_{t, \textrm{CLS}}, z_{t, 1}, z_{t, 2}, z_{t, 3}, ...$ to predict the current language instruction with $g_\psi$. An architecture figure can be found in Appendix C. 


\section{Experiments}
In this section we detail our experimental setup and empirical results. In particular, we investigate the benefits of instruction modeling for planning in limited data regimes. We seek to answer the following questions: How effective is instruction modeling loss? How does instruction modeling scale with both data and instruction annotations? What architecture choices are important? And finally, when is instruction modeling a fruitful objective?


\begin{figure*}[t]
\centering
\includegraphics[width=0.95\textwidth]{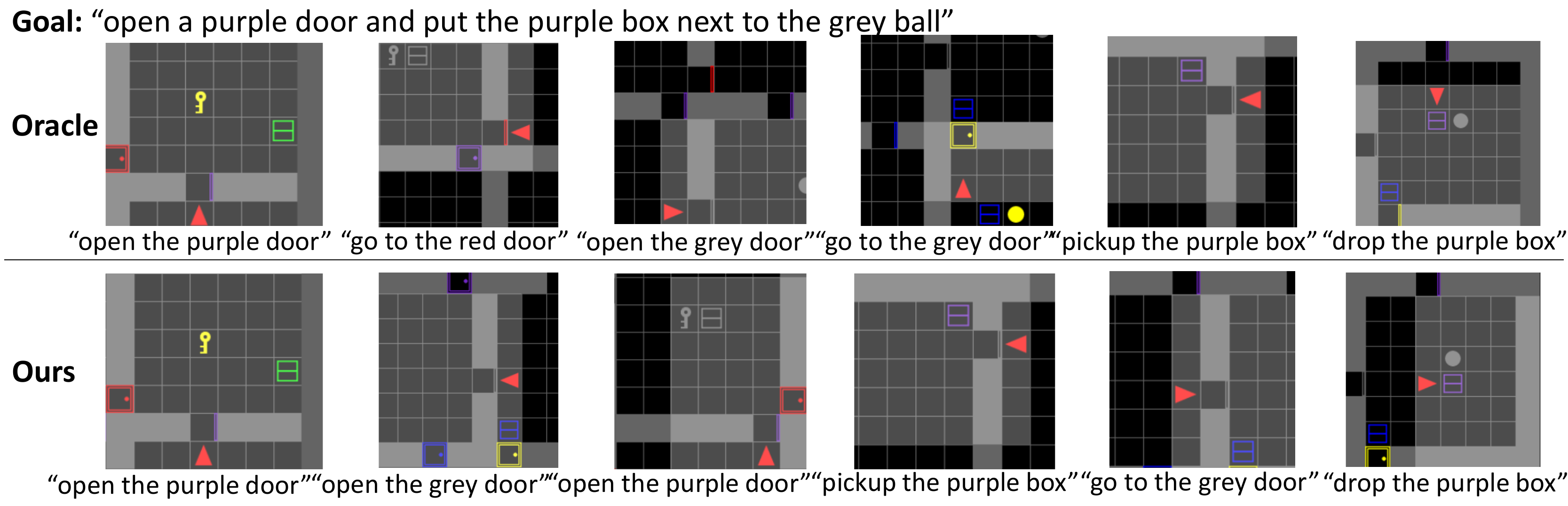}
\caption{Snapshots of a rollouts from an oracle agent and our trained agents on the same unseen task in BabyAI. Our agent is able to predict instructions, given below each image, with high fidelity. Our learned agent employs a different strategy but still completes the task exhibiting strong generalization.}
\label{fig:rollout}
\end{figure*}

\subsection{Environments}
To evaluate the effectiveness of instruction prediction at enabling long-horizon planning and generalization, we test our method on BabyAI \citep{babyai_iclr19} and the Crafting Environment from \citet{chen2021ask} which both provide coarse instructions. They cover challenges in partial observability, human generated text, and more.
We later examine the ALFRED environment to understand where instruction prediction is useful. Full model hyperparameters can be found in the Appendix.

\textbf{BabyAI:} Agents must navigate partially observable grid-worlds to complete arbitrarily complex goals specified through procedurally generated language such as moving objects, opening locked doors, and more. Agents are evaluated on their ability to complete unseen missions in unseen environment configurations. We modify the BabyAI oracle agent to output language instructions based on its planning logic. We focus our experiments on the hardest environment, BossLevel, and up to $10\%$ of the million demos in \citet{babyai_iclr19}. Because of partial observability, we employ a transformer sequence model as described in Section \ref{sec:instr_pred} with the same encoder from \citet{babyai_iclr19}. The language goals from the environment are tokenized and fed as additional inputs to the policies. We evaluate on five hundred unseen tasks.


\textbf{Crafting:} This environment from \citet{chen2021ask} tests how well an agent can generalize to new tasks using instructions collected from humans. The original dataset contains around 5.5k trajectories with human instruction labels. Each task is specified by a specific goal item the agent should craft, encoded via language. The agent must complete from one to five independent steps to obtain the final item. As this environment is fully observed, we employ the Vision Transformer based model described in Section \ref{sec:instr_pred} with the benchmark's original state encoder.



\subsection{Baselines}
We compare the effectiveness of our instruction modeling auxiliary loss to a number of baselines. The text in parenthesis indicates how we refer to the method in Tables \ref{tab:data}, \ref{tab:lang_amount}, \ref{tab:babyai_difficulty}, and \ref{tab:crafting_difficulty}. None of our models are pretrained, though we explore this and more additional baselines in the Appendix.
\begin{enumerate}
    \item \textbf{Original Architecture (Orig)}: The original state of the art model architectures proposed for each environment. The crafting environment uses a language-instruction hierarchy. In BabyAI, we use convolutions and FiLM layers as in \citet{babyai_iclr19}.
    \item \textbf{Transformer (Xformer)}: Our transformer based models without any auxiliary objectives to determine the effectiveness of our architectures.
    \item \textbf{Transformer Hierarchy (Hierarchy)}: 
    A high-level model outputs instructions for a low level executor for comparison to to hierarchical approaches.
    \item \textbf{Transformer with Forward Prediction (Forward)}: Instead of predicting instructions, we use the decoder to predict future actions. This baseline demonstrates the importance of using grounded information.
    \item \textbf{Transformer with ATC (ATC)}: Our transformer model with the active temporal contrast (ATC) self-supervised objective proposed in \citet{stooke2021decoupling}. This compares vision and instruction based representation learning. 
    \item \textbf{Transformer with Lang (Lang)}: Our transformer based models with just instruction prediction loss.
    \item \textbf{Transformer with ATC and Lang (Lang + ATC)}: Our transformer based models with both instruction modeling and constrastive auxiliary losses.
\end{enumerate}



\begin{table*}[]

\centering
\begin{tabular}{l|l|lllllll}
Env                                   & Demos & Orig         & Xformer      & Hierarchy    & ATC          & Forward      & Lang         & ATC+Lang     \\ \hline
\multirow{4}{1.3cm}{BabyAI BossLevel} & 100k  & 38.8       & 48.4         & 41.2         & 62.4         & 43.5          & \textbf{78.6} & 73.6    \\
                                      & 50k   & 35.3$\pm$0.1 & 40.2$\pm$2.2 & 36.8$\pm$3.5 & 45.8$\pm$0.6  & 37.0$\pm$0.6 & \textbf{70.3$\pm$1.3} & 64.3$\pm$0.5 \\
                                      & 25k   & 32.3$\pm$2.4 & 39.9$\pm$0.5 & 37.2$\pm$3.0 & 37.1$\pm$1.1 & 38.9$\pm$0.7 & \textbf{55.4$\pm$7.0} & \textbf{56.0$\pm$3.0} \\
                                      & 12.5k & 29.9$\pm$0.9 & 37.3$\pm$0.1 & 36.4$\pm$2.6 & 38.4$\pm$1.4 & 36.0$\pm$0.4 & \textbf{39.4$\pm$1.0} & 38.6$\pm$0.6 \\ \hline
\multirow{4}{1.3cm}{Crafting}         & 5k    & 9.4$\pm$1.1  & 75.8$\pm$3.6 & 63.2$\pm$10  & 78.8$\pm$3.1 & 77.9$\pm$4.3 & 75.9$\pm$2.4 & \textbf{80.3$\pm$2.5} \\
                                      & 3.3k  & 9.3$\pm$0.4  & 74.5$\pm$3.3 & 59.9$\pm$11  & \textbf{75.7$\pm$1.0} & 74.5$\pm$4.9 & 74.5$\pm$2.8 & \textbf{76.0$\pm$2.8}   \\
                                      & 2.2k  & 4.9$\pm$1.0  & 69.4$\pm$4.9 & 56.5$\pm$9.9 & 73.9$\pm$2.1 & 73.6$\pm$3.2 & 75.2$\pm$4.4 & \textbf{78.2$\pm$4.6} \\
                                      & 1.1k  & 1.7$\pm$0.8  & 70.1$\pm$3.8 & 39.4$\pm$3.8 & 70.1$\pm$3.7 & 73.0$\pm$3.8 & \textbf{74.8$\pm$2.6} & 71.4$\pm$2.9 \\ \hline
\multirow{1}{1.3cm}{ALFRED}           & 42k    & --         & 28.3 $\pm$ 1.0 & --           & --           & --           & 28.5$\pm$1.0        & --
\end{tabular}
\caption{Success rates (in \%) of all methods for varying numbers of demonstrations. The best method(s) is bolded, and the included range denotes the standard deviation (2 seeds for BabyAI and ALFRED, 4 for Crafting).}
\label{tab:data}
\end{table*}

\subsection{How Effective Is Instruction Prediction?}
Our main experimental results can be found in Table \ref{tab:data}, where we compare the performance of all methods on both environments with three differing dataset sizes. We find that for all environments and dataset sizes our instruction modeling objective improves or has no effect in the worst case. In BabyAI, we achieve a 70\% success rate on the hardest level with fifty thousand demonstrations and instructions. For comparison, it is worth noting that the original BabyAI implementation \citep{babyai_iclr19} achieved a success rate of 77\% with one million demonstrations. In the crafting environment, using instruction modeling boosts the success rate by about 5\% or more in the 1.1k and 2.2k demonstration setting. To our knowledge our results are state of art in this environment, exceeding the reported 69\% success rate on unseen tasks in \citet{chen2021ask} where RL is additionally used. We also find that the model is able to accurately predict instructions (Figure \ref{fig:rollout}), which appears to be correlated with performance. \textcolor{black}{See the Appendix for analysis of the language outputs of the models.}

Visual representation learning  was not as fruitful as language based representation learning overall. The combination of ATC and instruction modeling was unfortunately not constructive in all scenarios: it performed better in some instances and worse than just language loss in others. This is consistent with results found in \citet{chen2021empirical} that show that observation based auxiliary objectives often yield mixed results in the imitation learning setting. We find that our hierarchical implementations do not perform very well in comparison to plain transformer models. This is likely because with only a few demonstrations high-level language policies are likely to output incorrect instructions for unseen tasks leading low-level instruction conditioned policies to output sub-optimal actions. More analysis of the hierarchical baselines is in the Appendix. Finally, we see that the forward prediction used in the forward baseline hardly contributes to performance, indicating that grounded instructions do more than just combat compounding errors.



\begin{figure}
\centering
\includegraphics[width=0.98\columnwidth]{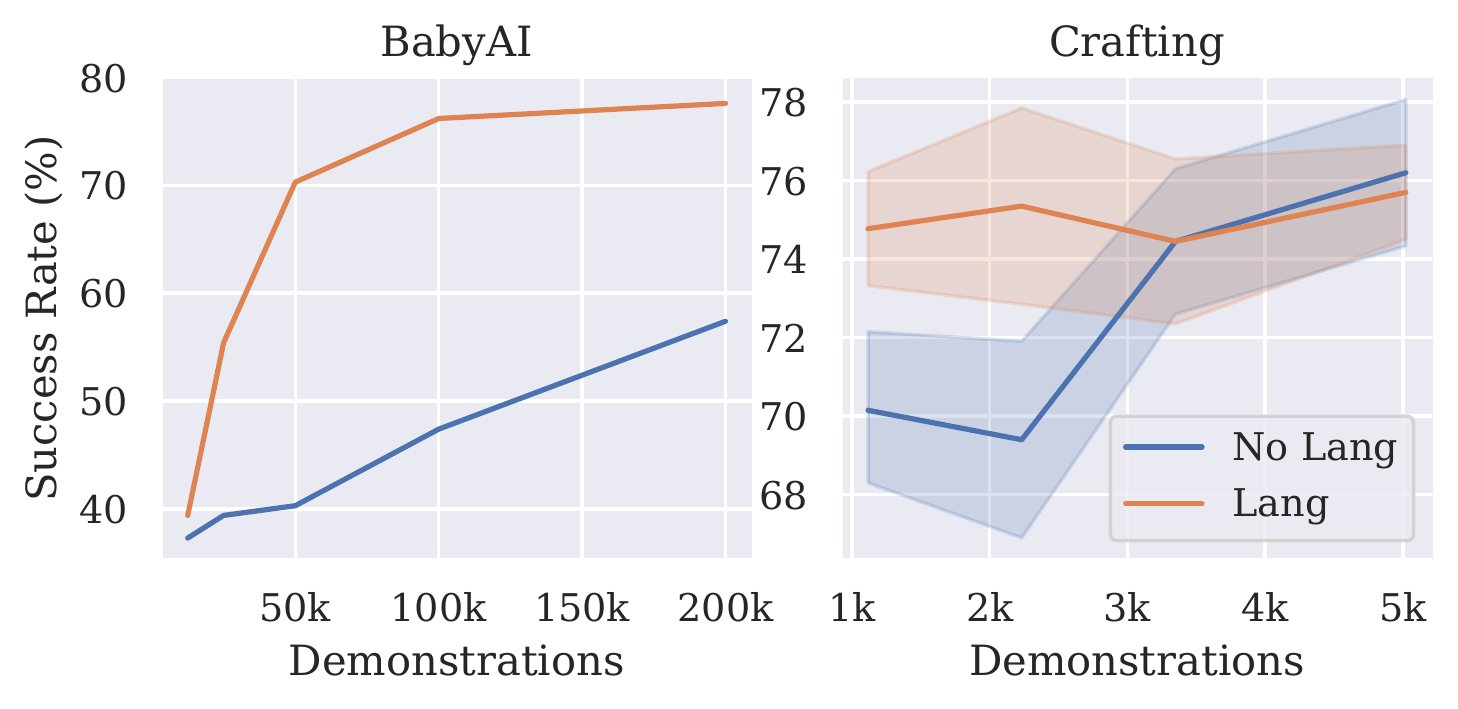}

\caption{Data scaling with and without instrucutions.}
\label{fig:scaling}
\end{figure}




\subsection{How Does Instruction Prediction Scale with Data and Annotations?} 
Overall, we find that instruction modeling reduces the amount of data required for policies to begin to generalize well. This is particularly evident in the low to medium data regime. With too little data, agents are likely to overfit quickly and only see a minor benefit from instruction prediction. With a significant amount of data instruction modeling may become unnecessary, and the policy can learn good representations from action labels alone. However, in between these regimes we find that instruction prediction reduces the amount of data needed to generalize by forcing the model to learn more ammendable representations to long-horizon planning. Figure \ref{fig:scaling} depicts how model performance changes with dataset size. In BabyAI, instruction modeling does not appear to significantly help with the smallest number of demos, however, after twelve and a half thousand demonstrations that we find that policy performance with language scales almost linearly with data before it experiences diminishing returns at two-hundred thousand demonstrations. Policies without language are unable to perform substantially better until we provide one hundred thousands demonstrations. This is not just because training with instructions helps overcome partial observability -- we show similar results on a fully observed version of BabyAI in the Appendix. The Crafting environment has only fourteen training tasks versus BabyAI's potentially infinite number, causing it to require fewer demos before performance saturates. Thus, we observe the opposite problem: instruction modeling helps when the policy is data constrained, and then is neutral when more data is introduced. In Section 4.6, we show that this saturation happens rather quickly for 5-step Crafting tasks, as there is only one in the training dataset.

A benefit of our loss-based approach is that it can easily be applied to mixed datasets that have only some instruction labels. To additionally study the scaling properties of our language prediction objective, we construct datasets in BabyAI where only half of the trajectories have paired instructions. Results can be found in Table \ref{tab:lang_amount}. Surprisingly, 
\textit{one is better off collecting 12.5k language annotations than collecting an additional 25k demonstrations} in the BabyAI environments. A similar statement can be made in the crafting environment for 1.1k demonstrations. This means that collecting instruction annotations is a feasible alternative to demonstrations.

\begin{table}

\centering

\begin{tabular}{l|lll}
\% w/ Instr & 0\%          & 50\%         & 100\%        \\ \hline
50k Demos       & 40.2$\pm$2.2 & 68.6$\pm$1.4 & 70.3$\pm$1.3 \\
25k Demos       & 39.9$\pm$0.5 & 50.3$\pm$1.3 & 55.4$\pm$7.0
\end{tabular}
\caption{We ablate the amount of demonstrations annotated with language instructions. Values are \% success rates with standard deviations.
}
\label{tab:lang_amount}
\end{table}

\begin{table}
      \centering
      
\begin{tabular}{l|l|ll}
Env  & Demos    & Goal   & Goal + Obs \\ \hline
BabyAI & 50K & 86.4\% & 92.4\%     \\
Crafting & 3.3K   & 53.8\% & 65.1 \%     \\
ALFRED & 42K & 96.9\% & 99.0\%    
\end{tabular}
\caption{Instruction prediction accuracies for models trained with and without access to observations. When instructions cannot be predicted from goals alone, better observation representations will be learned to help predict instructions.}
\label{tab:lang_pred}
\end{table}

\subsection{What Modeling Decisions Are Important?}
We ablate the use of our instruction decoder cross-attention masking in BabyAI. We find that the omission of the masking scheme leads to a 20\% drop in performance, from $70.3\pm1.3$\% to $50.1\pm12.1$\%. Without masking the language decoder has an easier time predicting an instruction as it can attend to observations from after the instruction finished, creating a disparity between train and test time, ultimately leading to lower quality representations. Overall, the transformer architecture appears to be critical to high performance, likely because of its good inductive bias for reasoning about objects and their interactions. This is especially evident in the Crafting environment. As stated in \citet{chen2021ask}, the imitation learning approaches with the original model were unable to achieve a meaningful success rate on any of the unseen tasks, whereas our baseline transformer achieves a success rate of around 70\%. Our architecture choice is also extremely parameter efficient as shown in the Appendix. 

\begin{table}[]

\centering
\resizebox{\columnwidth}{!}{
\begin{tabular}{l|ll|ll}
                 & \multicolumn{2}{l|}{50k Demos} & \multicolumn{2}{l}{25k Demos}  \\ \hline
Level            & Xformer        & Lang          & Xformer        & Lang           \\ \hline
GoTo      & 88.6$\pm$1.8   & \textbf{91.0$\pm$1.4}  & 77.6$\pm$2.2   & \textbf{82.3$\pm$3.3}  \\
SynthLoc   & 72.5$\pm$2.3   & \textbf{86.2$\pm$1.2 } & 60.1$\pm$0.5   & \textbf{69.4$\pm$1.6}  \\
BossLevel & 40.2$\pm$2.2   &\textbf{ 70.3$\pm$1.3}  & 39.9$\pm$0.5   & \textbf{55.4$\pm$7.0}  
\end{tabular}
}
\caption{Performance, in percent, of instruction prediction when varying the BabyAI level difficulty. The included range is the standard deviation.}
\label{tab:babyai_difficulty}
\end{table}

\begin{table}[]
\centering
\resizebox{\columnwidth}{!}{
\begin{tabular}{l|llll}
Demos         & Model    & 2 Steps         & 3 Steps         & 5 Steps         \\ \hline
\multirow{3}{*}{3.35k} & Xformer  & \textbf{98.1$\pm$0.4}  & 66.9$\pm$6.2  & \textbf{22.1$\pm$3.1} \\
                       & Lang     & 96.1$\pm$3.2  & \textbf{73.0$\pm$7.2}  & \textbf{19.3$\pm$4.4} \\
                       & Lang+ATC & 97.1$\pm$2.2  & \textbf{75.1$\pm$10.0} & 13.2$\pm$1.7 \\ \hline
\multirow{3}{*}{2.2k}  & Xformer  & 89.3$\pm$8.6  & 58.4$\pm$6.2  & \textbf{20.5$\pm$1.8}  \\
                       & Lang     & \textbf{96.1$\pm$0.7}  & \textbf{73.5$\pm$10.7} & 17.3$\pm$1.7  \\
                       & Lang+ATC & 93.9$\pm$2.7  & \textbf{78.3$\pm$13.9} & \textbf{19.8$\pm$2.7}  \\ \hline
\multirow{3}{*}{1.1k}  & Xformer  & 90.9$\pm$4.0  & 58.2$\pm$8.5  & 15.0$\pm$5.1  \\
                       & Lang     & \textbf{94.5$\pm$1.4}  & \textbf{76.0$\pm$8.4}  & 13.8$\pm$6.0  \\
                       & Lang+ATC & 89.5$\pm$3.2  & 65.2$\pm$10.7 & 11.6$\pm$2.8 
\end{tabular}
}
\caption{Difficulty comparison in the crafting environment. Steps indicate the number of steps required for the agent to craft the item. Performance is given in percent success rate with standard deviations.}
\label{tab:crafting_difficulty}
\end{table}

\subsection{When Is Instruction Prediction Useful?}
We hypothesize that instruction prediction is particularly useful for combinatorially complex, long horizon tasks. Many simple tasks, like ``open the door'' or ``grab a cup and put it in the coffee maker'' communicate all required steps and consequently stand to gain little from instruction modeling. Conversely, tasks in both environments we study do not communicate all required steps to agents. 
Thus, as task horizon and difficulty increase one would expect instruction modeling to be more important. In BabyAI we consider two additional levels -- GoTo, which only requires object localization, and SynthLoc which uses a subset of the BossLevel goals. Results in Table \ref{tab:babyai_difficulty} indicate that instruction modeling is indeed more important for harder tasks. 
The same trend approximately holds in the Crafting environment (Table \ref{tab:crafting_difficulty}). All policies are able to complete two steps tasks near or above 90\% success, but models with language prediction perform around 5\% better with fewer than 3.3k demonstrations. The difference in performance is ever greater for the three-step tasks, where instruction prediction boosts performance from around 58\% to closer to 75\% in most cases. Evaluations of the five-step tasks were noisy, which we attribute to the existence of only one five-step training task with which no model was able to adequately generalize. The takeaway from these observations is that instructions offer less training signal for combinatorially simple tasks, where reaching the goal requries only a few, obvious logical steps. Thus, we expect instruction modeling to not matter when instructions can easily be predicted from goals alone. 

To test this hypothesis, we train transformer models to predict instructions with and without access to observations in our primary environments and additionally in a modified version of the ALFRED benchmark \citep{ALFRED20}. Table \ref{tab:lang_pred} shows the token prediction accuracy 
of instructions using text goals alone versus text goals and observations. While token prediction accuracies are relatively high, particularly for BabyAI, accuracy differences of 5\% or more can make a large impact as instructions can share similar structure, but have a few critical tokens specifying objects. In the benchmarks where our method is impactful instructions cannot be as accurately predicted from text goals alone. However, in the ALFRED environment, which is largely based on visual understanding, instructions can easily predicted from just the goal. This indicates that while the visual complexity of ALFRED may be high, it does not pose significant challenges in logical understanding. 
Further analysis of the ALFRED benchmark is provided in the Appendix. In the future as tasks become more combinatorially complex, we expect instructions to provide a more critical modeling component.

\vspace{-0.1in}
\section{Conclusion}
We introduce an auxiliary objective that predicts language instructions for imitation learning and associated transformer based architectures. Our instruction modeling objective consistently improves generalization to unseen tasks with few demonstrations, and scales efficiently with instruction labels. We further analyze the domains where our method is successful, and make recommendations for when to apply it.

\bibliography{references}
\clearpage
\newpage
\appendix
\onecolumn
\section{Additional Experiments}
\subsection{Fully-Observed BabyAI}
In addition to the standard partially observed BabyAI environment, we created a fully observed version where the agent can view the entire world grid. In this fully observed setting we employ the same model architecture as in the Crafting environment, but with the hyperparameters from BabyAI. Results for a single seed on the BossLevel in the fully observed setting can be found in Table \ref{tab:fullyobs_babyai}, and look largely similar to those of the partially observed BabyAI environment.

\begin{table}[h]
\centering
\begin{tabular}{l|ll}
Demonstrations & XFormer & Lang   \\ \hline
50k            & 41.4\%  & 73.4\% \\
25k            & 39.8\%  & 56.2\% \\
12.5k          & 38.4\%  & 40.2\%

\end{tabular}
\caption{Performance, in percent of unseen tasks completed, of instruction prediction loss on a fully observed version of the BabyAI-BossLevel for a single seed.}
\label{tab:fullyobs_babyai}
\end{table}

\subsection{Analyzing Instruction Information and ALFRED}
\textcolor{black}{
In section 4.6 we find that instruction prediction is more useful as task difficulty increases. In this section, we try to measure how useful provided instructions are and additionally analyze instruction prediction in the ALFRED visual environment. Instruction prediction is likely to only provide a strong learning signal when they provide information not already available to the agent. Logically this makes sense: if all of the requisite information has been given to the agent in its task, instructions will add nothing new. For example, one goal from the ALFRED environment is ``put a watch on the table''. If the provided instructions for this task were ``pick up the watch'', ``go to the table'', and ``put down the watch'', the instructions would provide very little useful signal as all of their information was already conveyed by the goal.  
}

\textcolor{black}{
By default, ALFRED provides language instructions and goals as inputs to the agent. We remove the instructions from the input so they can be used for our auxiliary instruction prediction objective. Following the methodology and architecture choices of \cite{pashevich2021episodic}, we generate an additional 42K demonstrations in ALFRED and label them with vocabulary from the planner. Overall, we found that instruction prediction had little impact on performance across three seeds as seen in Table \ref{tab:alfred}. Based on these results, we assess why instruction prediction was not fruitful in ALFRED.
}
\begin{table}[]
\centering
\begin{tabular}{l|ll}
Success Measure       & Xformer        & Lang           \\ \hline
Task Success Rate     & 28.3$\pm$1.0\% & 28.5$\pm$0.7\% \\
Subgoal Success  Rate & 36.1$\pm$1.0\% & 36.0$\pm$0.8\%
\end{tabular}
\caption{We train models in the ALFRED environment with and without language prediction and evaluate their success on the ``seen'' validation set.}
\label{tab:alfred}
\end{table}


\textcolor{black}{
We can measure how much information instructions are able to provide to an agent by measuring how easy it is to predict them from just the goal. If instructions can easily be predicted from the goal alone, then they are unlikely to provide any additional learning signal to the agent. If the agent can only accurately predict instructions by observing the behavior in the demonstrations and the goal, then instructions prediction is more likely to encourage salient representation learning or logical reasoning about the task. For each benchmark, we take our transformer based architecture for instruction prediction and train only the language head to predict instructions with and without access to observations from the demonstrations. We report the best attained prediction accuracies in Table \ref{tab:lang_acc}.
}
\begin{table*}[]
\centering
\begin{tabular}{l|lll}
Model Inputs                 & BabyAI, 50K       & Crafting, 2.2k       & ALFRED, 42K  \\ \hline
Text Goal                    & 86.4\%            & 43.8\%               & 96.9\%            \\
Text Goal and Observation(s) & 92.4\%            & 49.9\%               & 99.0\%           
\end{tabular}
\caption{Instruction prediction accuracies for models trained with and without access to observations. When instructions can be easily predicted without access to observations, they likely provide little additional signal to the agent. The number next to the environment indicates the number of demonstrations used for training. The BabyAI level used was BossLevel, the hardest level.}
\label{tab:lang_acc}
\end{table*}
\textcolor{black}{
Our results indicate that the instructions in ALFRED can be nearly predicted perfectly from the text goals (96.9\%), indicating that the tasks are too easy and do not require instructions. This result has also been verified by the community. As of writing, one of the top models on the ALFRED leaderboard does not even use the language instructions as inputs. Conversely, we see lower accuracy overall and much larger gaps in accuracy for both BabyAI BossLevel (86.4\% to 92.4\%) and the Crafting environments (43.8\% to 49.9\%). We hope this result will drive the community to develop more logically challenging benchmarks with complex tasks where the scaling of instruction prediction can be further studied. Additionally, one can begin to assess the effectiveness of instruction prediction before using it by following this methodology.
}
\subsection{Additional Baselines}
\textcolor{black}{
We ran a additional baselines in the BabyAI environment.
\begin{enumerate}
    \item  \textbf{GPT Enc}: In order to demonstrate that instruction prediction is not just aiding in the agent's understanding of text goals, we construct a baseline that encodes the text goals in BabyAI using a pretrained GPT-2 Model before giving them to the agent. As the text-embeddings have been pretrained, this simulates the case where we have a maximal understanding of the text goal before interacting with the environment.
    \item \textbf{XFormer AC}: Our original architecture does not use previous actions as input due to the substantial increase in input tokens it causes. This baseline inputs both observation and action sequences into the transformer based model.
    \item \textbf{Goal Prediction}: Our architecture with language prediction, except instead of predicting the unseen instructions we predict the goal text that is used as input to the policy. This is a type of reconstruction objective in the text regime.
\end{enumerate}
We ran these additional baselines for two seeds. Results for these new baselines and \textit{XFormer} and \textit{Lang} can be found in Table \ref{tab:extras}. We find that encoding text goals with GPT does not lead to performance gas as large as language prediction. This indicates that our instruction prediction helps with learning good representations for planning, and not just language understanding. The transformer with action inputs (\textit{XFormer AC}) does not perform better than the regular transformer and in fact performs slightly worse, indicating that action inputs are not an important modeling component in the imitation domain and may just make learning harder by adding additional modalities and doubling sequence length. This is different than results found in the Offline RL setting in \cite{chen2021decision}, which makes sense as rewards often depend on both states and actions. Moreover, inverse models in the discrete action spaces in BabyAI are relatively easy to learn. Previous works with transformers in imitation \cite{pashevich2021episodic} have also found that conditioning on entire action sequences leads to a degradation of performance as policies can more easily overfit. 
}
\begin{table*}[h]
\centering
\begin{tabular}{l|lllll}
Demos & XFormer        & XFormer AC     & GPT Enc        & Goal Pred      & Lang           \\ \hline
50K   & 40.2 $\pm$ 2.2 & 37.4 $\pm$ 0.4 & 47.6 $\pm$ 0.4 & 43.5 $\pm$ 1.5 & 70.3 $\pm$ 1.3 \\
25K   & 39.9 $\pm$ 0.5 & 37.0 $\pm$ 0.3 & 37.6 $\pm$ 0.3 & 39.6 $\pm$ 1.9 & 55.4 $\pm$ 7.0
\end{tabular}
\caption{Results of additional baselines on the BabyAI Boss Level. The table gives success rates in \% on 500 unseen levels.}
\label{tab:extras}
\end{table*}

\subsection{Extended Training}
Due to space constraints, we have included results training on more demonstrations for BabyAI (100k and 200k demos) and Crafting (5k demos) here. These were already included in Figure \ref{fig:scaling}. In BabyAI, we were only able to run one seed for each model as they took far longer to converge with more data.

\begin{table*}[h]
\centering
\begin{tabular}{l|lllll}
BabyAI Demos   & 12.5k        & 25k          & 50k          & 100k         & 200k \\ \hline
XFormer        & 37.3$\pm$0.1 & 39.9$\pm$0.5 & 40.2$\pm$2.2 & 48.4         & 68.7 \\
Lang           & 39.4$\pm$1.0 & 55.4$\pm$7.0 & 70.3$\pm$1.3 & 78.6         & 81.8 \\ \hline \hline
Crafting Demos & 1.1k         & 2.2k         & 3.3k         & 5k           &      \\ \hline
XFormer        & 70.1$\pm$3.8 & 69.4$\pm$4.9 & 74.5$\pm$3.3 & 76.5$\pm$3.9 &      \\
Lang           & 74.8$\pm$2.6 & 75.2$\pm$4.4 & 74.5$\pm$2.8 & 74.5$\pm$3.3 &     
\end{tabular}
\caption{Results for training with more data in Crafting and BabyAI. }
\label{tab:extended}
\end{table*}

\section{Hierarchical Baselines}
Here we provide further details on our hierarchical baselines. The two prior works relevant on hierarchical language most relevant to our investigations are \cite{chen2021ask} and \cite{hu2019hierarchical}. Both of these works learn Markovian, or nearly-Markovian models. 

In the crafting environment \cite{chen2021ask}, the authors train a high-level RNN to output the current language instruction. They then condition their low-level policy on the latent representation fed to the RNN that predicts instructions. While they show latent condition to be effective, transformers purposefully avoid encoding entire streams of data into a single vector, and instead operate on a token level. Thus, we found it impractical to attempt this approach with our significantly more effective transformer models. 

The environment in \cite{hu2019hierarchical} is a partially observed multi-player strategy game. As alluded to in the related work, this environment has multiple viable strategies and is thus distinct from the oracle imitation learning we mostly consider. Though the environment is partially observed, the authors do not train a sequence model. Instead, they concatenate command data from previous time-steps to the model input. As information from the very beginning of the trajectory is necessary for some BabyAI tasks, we found it impractical to scale this concatenation based approach. In \cite{hu2019hierarchical}, a discriminative high level policy is trained to select an instruction from a fixed set. A low-level is trained with ground-truth human labeled instructions to output actions. In their strategy game, hierarchical approaches perform very well, unlike in our experiments where hierarchical models do not perform the best. One explanation for this comes from the nature of the strategy game environment. The distribution of optimal actions from a given state may be multi-modal, as different strategies may dictate different actions from the same state. Conditioning on an instruction would remove this multi-modality. Below we describe the hierarchical approaches we tried. For all approaches we used the same architectures as detailed in Section 3.

\textbf{Fully-Observed Setting.}
In the fully observed setting, we adopt a similar strategy to \cite{hu2019hierarchical}. A high-level policy takes as input the observation $o$ and goal $g$ and predicts the current instruction $x^{(i)}$.  We train a low-level policy that predicts actions from the current instruction, goal, and observation $\pi(a_t|o_t,x^{(i)},g)$. At test time, the high-level auto-regressively generates instructions that are then given to the low level.

\textbf{Partially-Observed Setting.}
Unfortunately, we find that there is no clear cut way to train a hierarchical model using only transformers where instructions and actions operate at different time scales. Here are the methods we tried:
\begin{enumerate}
    \item \textbf{Sequences for Each Instruction.} We take each trajectory take slices of it up until the completion of each instruction. Our high-level model predicts only the language instruction corresponding to that trajectory slice. This essentially means that the high-level predicts one instruction conditioned on all the history before the instruction. The same sequences are used to train the low-level, conditioned on the single instruction. The low-level policy can be written as $\pi(a_t|o_1, ..., o_t, g, x^{(i)}$. Because training sequence models with only individual losses is very inefficient, we train the low-level model to output the correct actions at all points in time corresponding to the instruction it is conditioned on.
    \item \textbf{All Instructions}. Instead of conditioning a policy on a single instruction, we train the high level policy to output all of the instructions for the entire task. This is especially challenging at the beginning of an episode when there are few frames. The low-level policy is then conditioned on the entire sequence of instructions and can be written as $\pi(a_t|o_1, ..., o_t, g, x^{(1)}, ..., x^{(n)})$. As this performs relatively well, we hypothesize that the model learns to ignore instructions far in the future when deciding which actions to take at the current timestep.
    \item \textbf{All Instructions, Aggressive Mask}. This is the same as the above, except we use an aggressive masking scheme when training the high level that only allows plans to be predicted from observations strictly preceding the time frame of the current instruction.
\end{enumerate}
In Table \ref{tab:hier} we give results for each of the sequence-style hierarchical approaches we tried on two seeds. In Table \ref{tab:data} we report the accuracy for the All Instructions method which we found to perform best. Other potential methods could include providing an encoding of the current instruction after each timestep to the transformer. However, such approaches would either employ an RNN or bag-of-words style model to generate the encoding and not be purely transformer based like the rest of our models.

\begin{table*}[]
\centering
\begin{tabular}{l|lll}
Demonstrations & Seq for Each Instr & All Instr       & All Instr, Aggressive Mask \\ \hline
50k            & 27.1$\pm$2.9\%     & 36.8$\pm$3.5\%  & 33.3$\pm$2.8\%            \\
25k            & 25.5$\pm$3.7\%     & 37.2$\pm$3.0 \% & 32.4$\pm$2.3                        
\end{tabular}
\caption{Results for hierarchical configurations we tried.}
\label{tab:hier}
\end{table*}

\section{Additional Figures}
\begin{figure}
\centering
\includegraphics[width=0.5\textwidth]{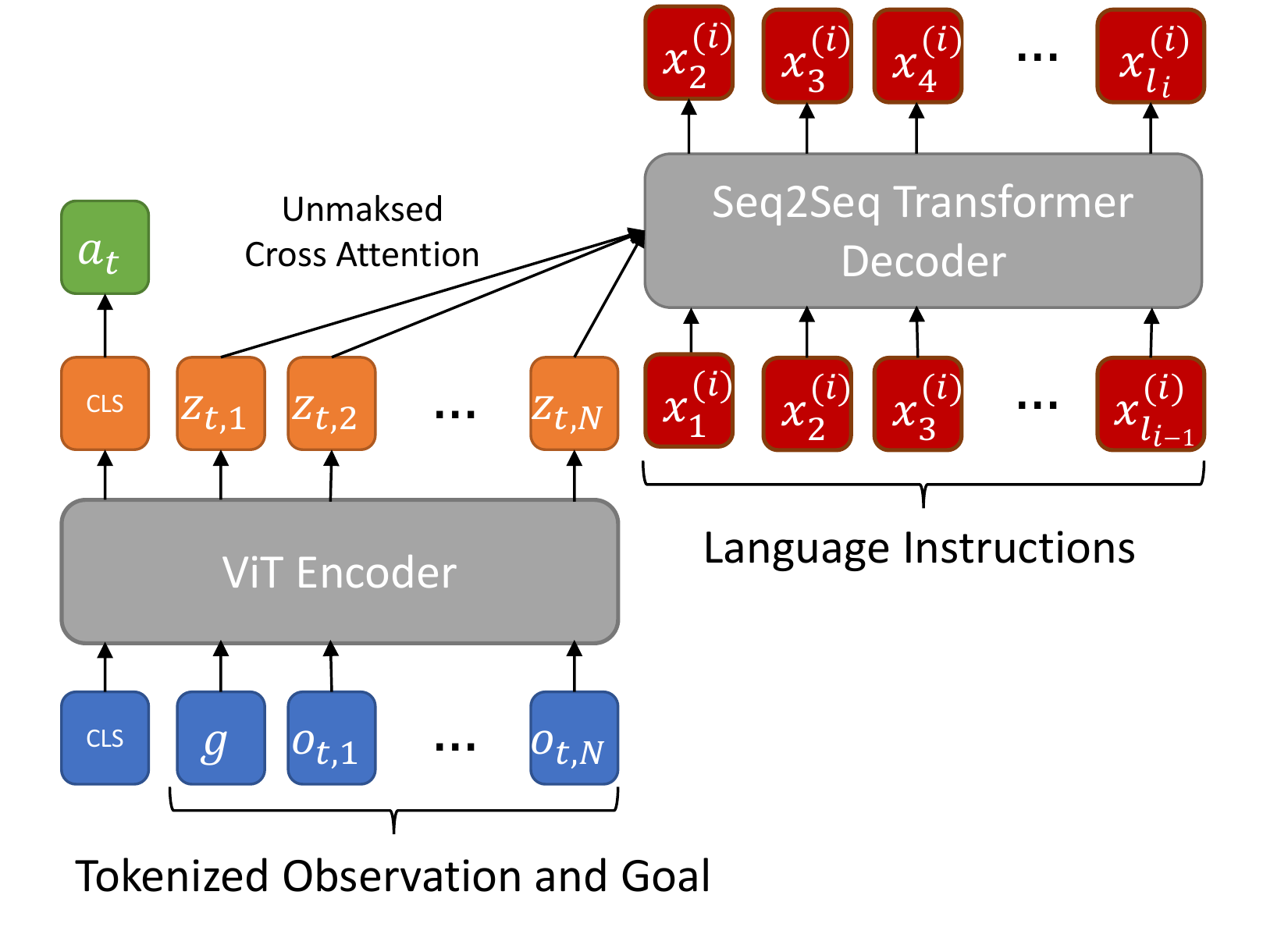}
\vspace{-0.1in}
\caption{Architecture figure for the fully observed case. Observations are tokenized and then input to a Vision-Transformer like encoder.}
\label{fig:vit_arch}
\end{figure}

See Figure \ref{fig:vit_arch} for the Fully Observed ViT-based architecture.



\section{Language Analysis}
Below we include metrics on the performance of the instruction prediction component of our models.
\begin{table*}[h]
\centering
\begin{tabular}{l|lllll}
BabyAI Demos        & 12.5k  & 25k    & 50k    & 100k   & 200k   \\
Prediction Accuracy & 88.4\% & 89.8\% & 91.2\% & 92.0\% & 92.2\% \\
BLEU Score          & 0.753  & 0.783  & 0.808  & 0.823  & 0.827  \\ \hline
Crafting Demos      & 1.1k   & 2.2k   & 3.3k   & 5k     &        \\
Prediction Accuracy & 60.6\% & 63.6\% & 65.1\% & 65.9\% &        \\
BLEU Score          & 0.353  & 0.396  & 0.408  & 0.421  &       
\end{tabular}
\caption{Language modeling metrics for the instruction prediction models.}
\label{tab:lang}
\end{table*}
As is in Table \ref{tab:lang}, the instruction prediction models attain high accuracy and BLEU scores which tend to increase with more data. We find that the synthetically generated instructions in babyAI are much easier to model than the real human language in the crafter benchmark, which tends to be more multi-modal. Below we also include example outputs of our language model trained with varying amounts on data. Trajectories are fed into the model, and it is asked to reproduce the instructions associated with them as during training. For the BabyAI model, causal masking is still used. Overall, the model trained with more data more accurately predicts the instructions. We did not use any beam-search style methods to produce the outputs.

\begin{table*}[h]
\centering
\begin{tabular}{p{0.22\linewidth} | p{0.22\linewidth} | p{0.22\linewidth} | p{0.22\linewidth}}
Ground Truth & 1.1K Demos & 3.3K Demos & 5K Demos \\ \hline
go to key grab it go to door and open door  & grab to the and it go to door and open door  & grab to key and it go to door and open door  & grab to key \\
move to wood plank press craft to make plank  & make to wood plank crafting craft  & craft to wood plank press craft  & craft to wood plank and craft wood make wood \\
move to stone pickaxe press craft i  & go to stone pickaxe and craft  & craft to stone pickaxe  & craft to stone pickaxe press craft \\
move to switch press toggle switch  & toggle to switch and open switch  & toggle to the and toggle switch  & toggle to switch and toggle switch \\
move to cobblestone stash press mine  & go to the stash  & mine to cobblestone stash and mine  & go to cobblestone stash and mine \\
open the door with the key  & open door door  & go the door  & go door door \\
craft at the wood plank table  & go at the wood plank bench  & craft wood the wood plank table  & go a the wood plank table \\
move to pickaxe press grab  & go to pickaxe and grab  & grab to pickaxe press grab  & grab to pickaxe press grab \\
move to tree press mine  & walk to tree and mine to & go to the press mine  & mine right the press mine \\
chop the tree  & mine down tree  & go the tree  & go down tree \\
\end{tabular}
\caption{Generated instructions for validation states on the crafter benchmark}
\label{tab:ayh_gen}
\end{table*}

\section{Hyperparameters}
Here we include all hyper-parameters we used. We used pytorch for our experiments. Our implementation for the partially observed sequence models was based on MinGPT by Andrej Karpathy. Our implementation of the Vision Transformer for fully observed environments was based on \cite{rw2019timm}. We determined parameters for ATC by testing different frame skip values in both environments. In the crafting environment we tested using $\lambda$ coefficients of 0.5 and 0.25 and found 0.25 to perform better. We also tested using weight decay and dropout in BabyAI with 50k demonstrations and found it to have no significant impact and thus did not use it for the other experiments. When evaluating models we found those for the Crafting environment to perform better in Pytorch ``train'' mode, meaning with dropout on, while those for BabyAI worked better in ``eval'' mode. We ran our experiments on NVIDIA GTX 1080 Ti GPUs. In BabyAI we train all models for two seeds and in the crafting environment we train all models for four. For 100k or more demonstrations in BabyAI, we train only one seed as is done in \cite{babyai_iclr19}. This was done because we found that training transformers until convergence on BabyAI takes around 4 days on GPU with 100k demonstrations, and even longer for 200k. 

\begin{table}[h!]
\centering
\begin{tabular}{l|ll}
Hyperparameter          & BabyAI           & Crafting         \\ \hline
Encoder Blocks          & 4                & 4                \\
Decoder Blocks          & 1                & 1                \\
Embedding Dim           & 128              & 128              \\
MLP Size                & 256              & 256              \\
Dropout                 & 0                & 0.1              \\
policy $\pi_\phi$       & Dense Layer      & Dense Layer      \\
Batch Size              & 32               & 64               \\
Training Steps (Table 2)& 1 million        & 300k             \\
Training Steps (Larger data) & 3 million & 350k             \\
Optimizer               & Adam             & AdamW            \\
Optimizer Epsilon       & $1\times10^{-8}$ & $1\times10^{-8}$ \\
Learning Rate           & 0.0001           & 0.0001           \\
Weight Decay            & 0.0              & 0.05             \\
Grad Norm Clip          & N/A              & 1                \\
$\lambda_{\text{lang}}$ & 0.7              & 0.25             \\
$\lambda_{\text{ATC}}$  & 0.7              & 0.25             \\
EMA $\tau$              & 0.01             & 0.01             \\
EMA update freq         & 1                & 1                \\
ATC Frame Skip          & 3                & 1               
\end{tabular}
\caption{Hyperparameters}
\label{tab:hparams}
\end{table}

\begin{table}[]
\centering
    \begin{tabular}{l|lll}
Model           & Base    & Lang   & ATC \\ \hline
Boisvert et al. & 1052297 & -      & -     \\
GPT-Like & 661385  & 253056 & 98816 \\
Chen et al. & 1379980 & 218083 & -     \\
ViT-like & 579177  & 259712 & 98816
\end{tabular}
\caption{Model Parameter counts by component. The hierarchical transformer baselines roughly double the parameter counts of their respective models.}
\label{tab:params}
\end{table}

\clearpage
\onecolumn
\begin{longtable}[c]{c|c|c|c}

Ground Truth & 12.5K Demos & 50K Demos & 100K Demos \\ \hline \endfirsthead

Ground Truth & 12.5K Demos & 50K Demos & 100K Demos \\ \hline \endhead

\endfoot
\endlastfoot

open the grey door  & open the grey door  & open the grey door  & open the grey door \\
 open the yellow door  &  open the yellow door  &  open the yellow door  &  open the yellow door \\
 go to the red ball  &  open to the red ball  &  open to the red ball  &  open to the red ball \\
 open the purple door  &  open the yellow door  &  open the yellow door  &  open the yellow door \\
 open the yellow door  &  open the purple door  &  open the yellow door  &  open the yellow door \\
 pick up the red key  &  open up the red key  &  open up the red key  &  open up the red key \\ \hline
 
pick up the purple key  & pick up the blue key  & pick up the purple key  & pick up the purple key \\
 open the yellow door  &  open the yellow door  &  open the yellow door  &  open the yellow door \\
 move the blue key  &  open the blue key  &  open the blue key  &  open the blue key \\
 open the purple door  &  open the grey door  &  open the purple door  &  open the purple door \\
 go to the green door  &  open to the green door  &  open to the green door  &  open to the green door \\
 go to the blue key  &  open to the blue key  &  go to the blue key  &  go to the blue key \\ \hline
 
 open the yellow door  & open the yellow door  & open the yellow door  & open the yellow door \\
 open the blue door  &  open the blue door  &  open the blue door  &  open the blue door \\
 open the yellow door  &  open the red door  &  open the yellow door  &  open the green door \\
 open the green door  &  open the red door  &  open the green door  &  open the green door \\
 open the yellow door  &  open the red door  &  open the yellow door  &  open the yellow door \\
 open the blue door  &  open the red door  &  pick the blue door  &  open the green door \\
 open the green door  &  open the blue door  &  pick the blue door  &  open the green door \\
 open the blue door  &  pick the blue door  &  pick the blue door  &  pick the blue door \\
 pick up the purple box  &  pick up the purple box  &  pick up the purple box  &  pick up the purple box \\
 go to the yellow key  &  go to the yellow key  &  go to the yellow key  &  go to the yellow key \\
 drop the purple box  &  drop the purple box  &  drop the purple box  &  drop the purple box \\
 pick up the grey key  &  pick up the grey key  &  pick up the grey key  &  pick up the grey key \\ \hline
 
open the red door  & open the red door  & open the red door  & open the red door \\
 open the green door  &  open the green door  &  open the green door  &  open the green door \\
 open the green door  &  open the blue door  &  open the purple door  &  open the purple door \\
 go to the purple box  &  open to the purple box  &  open to the purple box  &  open to the purple box \\
 pick up the blue box  &  pick up the red box  &  pick up the blue box  &  pick up the blue box \\
 go to the green ball  &  go to the green ball  &  go to the green ball  &  go to the green ball \\
 drop the blue box  &  drop the grey box  &  drop the blue box  &  drop the blue box \\ \hline
 
open the blue door  & open the blue door  & open the blue door  & open the blue door \\
 open the purple door  &  open the purple door  &  pick the purple door  &  pick the purple door \\
 open the yellow door  &  pick the blue door  &  pick the green door  &  pick the green door \\
 open the green door  &  pick the blue door  &  pick the green door  &  pick the green door \\
 pick up the yellow box  &  pick up the green box  &  pick up the purple box  &  pick up the purple box \\
 go to the grey ball  &  go to the green ball  &  go to the grey ball  &  go to the grey ball \\
 drop the yellow box  &  drop the green box  &  drop the yellow box  &  drop the yellow box \\
 open the green door  &  go the green door  &  go the yellow door  &  open the green door \\
 pick up the green box  &  move up the green box  &  open up the yellow box  &  open up the green box \\
 go to the green door  &  go to the green door  &  go to the green door  &  go to the green ball \\
 drop the green box  &  drop the green box  &  drop the green box  &  drop the green box \\ \hline
 
open the purple door  & open the purple door  & open the purple door  & open the purple door \\
 open the red door  &  open the red door  &  open the red door  &  open the red door \\
 open the red door  &  open the grey door  &  open the red door  &  open the red door \\
 open the green door  &  open the grey door  &  pick the red door  &  open the red door \\
 pick up the green ball  &  pick up the green ball  &  pick up the green ball  &  pick up the green ball \\
 go to the blue key  &  go to the blue key  &  go to the blue key  &  go to the blue key \\
 go to the grey door  &  go to the grey key  &  go to the grey door  &  go to the grey door \\
 close the grey door  &  close the grey door  &  close the grey door  &  close the grey door \\
 open the grey door  &  open the grey door  &  open the grey door  &  open the grey door \\
 pick up the grey ball  &  pick up the yellow ball  &  pick up the blue ball  &  pick up the red ball \\
 go to the purple key  &  go to the blue key  &  go to the purple key  &  go to the purple key \\
 drop the grey ball  &  drop the grey ball  &  drop the grey ball  &  drop the grey ball \\ \hline
 
open the grey door  & open the grey door  & open the grey door  & open the grey door \\
 open the yellow door  &  open the yellow door  &  open the yellow door  &  open the yellow door \\
 open the grey door  &  open the blue door  &  open the yellow door  &  open the blue door \\
 pick up the grey key  &  open up the grey key  &  open up the grey key  &  open up the grey key \\
 go to the purple door  &  go to the grey door  &  go to the grey door  &  go to the yellow door \\
 drop the grey key  &  drop the grey key  &  drop the grey key  &  drop the grey key \\
 move the grey key  &  open the grey key  &  move the grey key  &  move the grey key \\
 open the purple door  &  open the purple door  &  open the purple door  &  open the purple door \\
 open the grey door  &  open the yellow door  &  open the yellow door  &  open the yellow door \\
 go to the grey ball  &  open to the grey ball  &  open to the grey ball  &  open to the grey ball \\ \hline
 
open the yellow door  & open the yellow door  & open the yellow door  & open the yellow door \\
 open the yellow door  &  open the yellow door  &  open the yellow door  &  open the yellow door \\
 open the purple door  &  open the grey door  &  open the blue door  &  open the blue door \\
 open the purple door  &  open the grey door  &  open the purple door  &  open the purple door \\
 move the yellow box  &  open the red key  &  open the blue box  &  open the purple key \\
 open the green door  &  open the green door  &  open the green door  &  open the green door \\
 pick up the green ball  &  pick up the green ball  &  move up the green ball  &  open up the green ball \\
 go to the green door  &  go to the red door  &  go to the green door  &  go to the green door \\
 drop the green ball  &  drop the green ball  &  drop the green ball  &  drop the green ball \\ \hline

\caption{Generated instructions on unseen levels of BabyAI BossLevel. We display eight examples from longer levels which have more complex goals and thus longer instructions. Each line of the table gives a subgoal, and the horizontal lines separate different tasks.}
\label{tab:babyai_gen}
\end{longtable}
\clearpage
\twocolumn


\end{document}